\documentclass[a4paper]{llncs}
\usepackage{graphicx, wrapfig, float, hyperref, xcolor}
\graphicspath{{images/}}

\usepackage{amsmath}
\usepackage{amssymb}
\usepackage{tabulary}

\usepackage{booktabs}

\makeatletter 
\renewcommand\@biblabel[1]{#1} 
\makeatother
\begin{document}
\title{Topological Data Analysis of Time Series Data for B2B Customer Relationship Management}

\titlerunning{TDA of Time Series Data for B2B CRM}

%
\author{Rodrigo Rivera-Castro\inst{1} \and
Polina Pilyugina\inst{1} \and
Alexander Pletnev \inst{1} \and
Ivan Maksimov \inst{1} \and
Wanyi Wyz  \inst{2} \and
Evgeny Burnaev\inst{1}}
\authorrunning{R. Rivera et al.}
%
\institute{Skolkovo Institute of Science and Technology, \email{rodrigo.riveracastro@skoltech.ru},\\
\and
Alibaba Cloud Intelligence Business Group}%

\maketitle              
\begin{abstract}
Topological Data Analysis (TDA) is a recent approach to analyze data sets from the perspective of their topological structure. Its use for time series data has been limited to the field of financial time series primarily and as a method for feature generation in machine learning applications. In this work, TDA is presented as a technique to gain additional understanding of the customers' loyalty for business-to-business customer relationship management. Increasing loyalty and strengthening relationships with key accounts remain an active topic of discussion both for researchers and managers. Using two public and two proprietary data sets of commercial data, this research shows that the technique enables analysts to better understand their customer base and identify prospective opportunities. In addition, the approach can be used as a clustering method to increase the accuracy of a predictive model for loyalty scoring. This work thus seeks to introduce TDA as a viable tool for data analysis to the quantitate marketing practitioner.

\end{abstract}

\section*{ORIGINALITY AND VALUE}\label{sec:originality}
This research presents a system for customer base analysis and demand forecasting developed for a leading provider of cloud computing. Validated with real data, the approach has yet to be deployed in production. The contributions cover the areas of data pre-processing of customer relationship management (CRM) data and customer demand prediction. The proposed system is suited for individuals with domain knowledge but limited understanding of machine learning methods. The contributions are the following: a) An industry case of customer base analysis and demand prediction for a major provider of cloud computing, b) an evaluation of three different models for customer segmentation, two of them represent an original work, c) a presentation of time series clustering methods for customer segmentation, d) an assessment of Topological Data Analysis techniques for CRM data, f) a novel and relevant data set from a B2B digital provider in the hospitality industry, g) for reproducibility purposes, an implementation and data set available for download\footnote{\url{https://github.com/rodrigorivera/imp19}}.

\section*{PROBLEM STATEMENT}
\begin{figure}[!htb]
\begin{center}
\includegraphics[width=0.6\columnwidth]{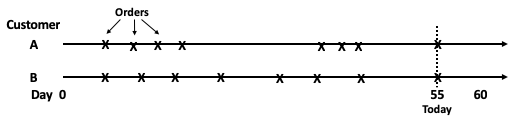} 
\end{center}
\caption{Two customers can share the same Recency, Frequency and Monetary scores. Yet, customer A is likely more alive than B}
\label{fig:general:customer_timelines}
\end{figure}

\begin{figure}[!htb]
\begin{center}
\includegraphics[width=0.5\columnwidth]{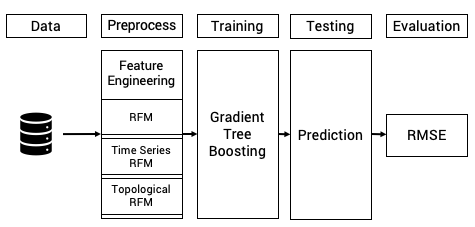} 
\end{center}
\caption{Proposed machine learning pipeline with 3 different variations of the RFM framework}
\label{fig:general:process_diagram_1}
\end{figure}

One of the world's largest cloud computing providers requires a better understanding of its customer base, in order to improve demand forecasting for its services. The individual customer demand amounts to millions of time series data to predict. Given the novelty of the cloud computing offerings and the flexibility it offers to customers, historic data is limited, seasonality hard to detect and historic records often non-representative. In summary, the data available at a customer level is limited and hard to work with. As a consequence, traditional forecasting techniques are largely ineffective. Further, popular heuristics such as the Recency, Frequency, Monetary framework can be misleading with two customers sharing the same score while being very different, as seen in \autoref{fig:general:customer_timelines}. At the same time, the provider is interested in obtaining a deeper understanding of its customer base to adjust its product offering and promotions while being able to generate reliable estimates for their future demand over multiple periods. Thus, rather than relying purely on traditional heuristics such as the Recency, Frequency, Monetary (RFM) framework to understand the customer base, this work proposes a machine learning pipeline consisting of three different methods depicted in \autoref{fig:general:process_diagram_1}.

\section*{RESEARCH ABSTRACT AND GOALS} \label{sec:research_abstract}
The objective of this research is to present two techniques for customer data segmentation and prediction accessible to non-technical business experts. They are motivated by the works of \cite{Zhang2015-wa} and \cite{Platzer2016-si} in CRM to provide model-based approaches emphasizing timing patters to predict future purchase activities. The use of novel machine learning methods is a promising area with little academic research and insufficient efforts to expose practitioners to them \cite{2017arXiv170905548R,Rivera2018}. In addition, over 40\% of analysts still use primarily traditional forecasting methods \cite{Chase2013}.
There are significant incentives to develop methods that can be easily adopted by quantitative marketeers. In forecasting, for discrepancies as low as 2\%, it is worth improving the accuracy of a forecast, \cite{Fleisch2003}. Yet, companies struggle hiring the adequate personnel to address these tasks. For example, by 2020, Vietnam is expected to face a shortage of over 500,000 employees with data science and analytics skills and over 80\% of the local workforce do not have the necessary skill set to fill this gap \cite{2017apec}. In Europe, a survey by \cite{2018esade} reported that over 70\% of surveyed businesses struggled hiring data science personnel and over 60\% are resorting to internal training to upgrade the skills of existing business analysts. This work seeks to alleviate this situation by presenting two customer segmentation techniques based on state-of-the-art methods that are both accurate as well as easy to communicate to decision-makers.
The research goal of this work is to propose a set of approaches for customer segmentation that can be adopted by business practitioners. For this purpose, the study poses the questions: 1) Although RFM is very attractive for practitioners because it only requires computing and monitoring three variables, is this framework sufficient to explain key aspects of customer behavior? 
2) How can RFM be extended to improve the accuracy of predictive models?  To achieve the research goal, two objectives have been assigned: a) To review the existing theory on enriching RFM to measure customer loyalty more accurately; b) To make a performance comparison between RFM and the two proposed techniques.
The object of research is the balance between accessibility and precision of methods for customer segmentation using time series clustering and topological data analysis within the industry. 
The subject of the research is customer segmentation combined with prediction of customer's next action.

\section*{LITERATURE REVIEW} \label{sec:literature_review}
\subsection*{Customer Loyalty}
The literature covering Customer Loyalty is vast and a thorough review is out of the scope of this work. This study narrows it down by focusing on the combination of existing frameworks for customer classification extended with machine learning methods. Examples of this are the combination of the Recency Frequency Monetary framework with other techniques and extending its scope can be seen in the recent work of \cite{Zaki2016-oj}. They combined the Net Promoter Score, a survey-based metric commonly used to predict customer satisfaction and repurchase intention, together with RFM; thus, giving additional meaning to NPS by adding a quantitative factor based on purchase history. This study follows the argument made by \cite{Wubben2008-yf} that the use of big data techniques must be used to update the customer loyalty measurement in organizations. Firms benefit from the use of sophisticated and advanced approaches. They help uncover patterns in customer data, which can be linked to business results \cite{doi:10.1108/JOSM-01-2013-0018}.

\subsection*{Recency Frequency Monetary}
Recency Frequency Monetary (RFM) is a managerial metric originated in database marketing. In its original form, it seeks to increase response rates by classifying customers into five equal groups based on aspects of their past behavior. As a result, a three-digit number is obtained. The lower the number, the higher the probability of customer churn \cite{doi:10.1177/1094670506293810}.

\subsection*{Time Series Clustering}
Clustering time-series data is a technique used in many areas to discover patterns. Broadly, clustering represents partitioning $n$ observations into $k$ clusters, where a cluster is characterized with the notions of homogeneity, the similarity of observations within a cluster, and separation, which is the dissimilarity of observations from different clusters. In the context of time series, \cite{Aghabozorgi2015} argues that their unique characteristics make them unsuitable to conventional clustering algorithms. In particular, the high dimensionality, very high feature correlation, and typically large amount of noise have been viewed detrimental to their performance. Further, \cite{Paparrizos2017-nw} highlights three main drawbacks in methods for time series clustering: (i) they cannot easily scale to large volumes of data, (ii) they are domain-specific or only work for specific data sets, and (iii) they are sensitive to outliers and noise. 

\subsection*{Topological Data Analysis}
Topological Data Analysis (TDA) is a recent field that emerged from a combination of various statistical, computational, and topological methods during the first decade of the century. It allows to find shape-like structures in the data and has proven to be a powerful exploratory approach for noisy and multi-dimensional data sets. For a detailed introduction, the reader is invited to consult \cite{chazal2017introduction}. \cite{Turner2019-bc} highlight that TDA is usually concerned with analyzing complex data with a complicated geometric or topological structure. It is possible to represent this structure with a family of topological spaces, a filtration, defined as $\{K_a\}_{a \in A \subset \mathbb{R}}$ if $K_a \subset K_b$ whenever $a \leq b$. The inclusion of $K_a \subset K_b$ induces a homomorphism between the homology groups $H_k(K_a)$ and $H_k(K_b)$. The persistent homology is an image of $H_k(K_a)$ in $H_k(K_b)$, it encodes the $k$-cycles in $K_a$ that are independent with respect to boundaries in $K_b$. Thus, $H_k(a,b):= \frac{Z_k(K_a)}{(B_k(K_b) \cap Z_k(K_a))}$ with $Z_k$ as the cycle group and $B_k$ as the boundary group, both subgroups of the $k$th chain group $C_k$ of $K$, a free Abelian group on its set of oriented $k$-simplices. The popular representations of persistent homology information are the barcode and the persistence diagram. A barcode is a collection of intervals [birth,death) each representing the birth and death values of a persistent homology class. This collection of intervals satisfies the condition that for every $a \leq b$, the number of intervals containing $[a,b)$ is $dim(H_k(a,b))$. A persistence diagram is the multi-set of points in the plane where each bar in the barcode is sent to the point with first coordinate, its birth time, and its second coordinate, its death time. After a filtration of topological spaces is built from the observations, a persistent homology is applied. This filtration can be summarized in terms of the evolution of the homology. Thus, a summary from a single complex object is created. A wide array of topological summaries can be computed directly from a persistence diagram or barcode. Each of these is a different expression of the persistent homology in the form of a topological summary statistic.

\section*{DATA SETS}\label{sec:dataset}
\textbf{Bimbo}: The bimbo data set contains 2000 retail points for baked goods produced by the Bimbo manufacturer with nine weeks of sale data. Each retail point operates independently from the manufacturer and although a contractual relationship exists among them, each week the manufacturer has to supply a different product amount. The data set can be found in Kaggle\footnote{\url{https://www.kaggle.com/c/grupo-bimbo-inventory-demand/data}}, a website for data science competitions.

\textbf{CDNOW}: The CDNow data set has been commonly used in the CRM literature. It contains historic records of a cohort comprising 23,570 individuals from their first purchase in the first quarter of 1997 up to the end of June 1998. The data set can be downloaded \footnote{\url{http://www.brucehardie.com/datasets/}}.

\textbf{B2B Hospitality Procurement}: This data set consists of 1900 customers of a business-to-business digital company in the goods procurement sector for the hospitality industry. The data set covers a six month period in 2018. The data set is being made public and available for download\footnote{\url{https://github.com/rodrigorivera/imp19}}.

\textbf{Cloud Computing Provider}: The data represents a small subset of the customer base. It contains observations with time stamps documenting whenever a customer has booked computing (CPU) time with the provider between 2017 and 2018, the duration and the type of product booked.
\section*{METHODS}
\label{sec:methods}

\textbf{RFM + Prediction}: In this method, RFM is applied to each of the four data sets. From RFM, three new attributes are obtained and they are embedded on each customer. They serve as new data features. This task can be seen as part of the data pre-processing phase in a machine learning task. The enriched data set is then fitted in a predictive model. This work selected gradient tree boosting as the method of choice due to its pervasiveness in the industry, its robustness and state-of-the-art results. The purpose of using 'RFM + Prediction' is to benchmark this results against those obtained with 'Time Series RFM' and 'Topological RFM'.

\textbf{Time Series RFM}: In this work, K-Shape by \cite{Paparrizos2017-nw}, a time-series clustering technique based on shape, is used. It is efficient, domain independent and comparable to state-of-the-art methods such as Dynamic Time Warping (DTW) with K-Means. The proposed method requires following phases: 1) As a first step, three time series were generated for each user. They correspond to the Recency, Frequency and Monetary values. Thus, instead of having a point value for Recency, a time series is provided. 2) Once the time series have been prepared, as a second step, they are used as an input in K-shape. Here, three instances are started. Each of them with four clusters. The number of clusters was decided through trial and error by visually inspecting the generated clusters. 3) As a third step, the results from each of the three K-shape instances are embedded into the data. With the extended data set, a gradient tree boosting model is fitted.

\textbf{Topological RFM}: This work is inspired by \cite{Seversky2016-dz} and \cite{Gidea2017-em}. The architecture proposed for 'Topological RFM' is divided into five different steps. 1) As a first step, three time series are generated for Recency, Frequency and Monetary respectively. This step is akin to the first step in 'Time Series RFM'. 2) The time series are sliced using sliding windows. The objective is to generate delay embeddings that can be projected as a point cloud. 3) Once the three point clouds have been obtained, Rips filtration, a popular algorithm in TDA, is used, with the objective of obtaining death and birth complexes. 4) As a fourth step, barcode diagrams are generated for both 0- and 1- dimensional homologies. They help to visualize the birth-death filtered complexes. The focus is on the 1-dimensional homologies (loops). 5) As a final step, a clustering is done using K-means based on features extracted from the barcodes. The number of clusters for each of Recency, Frequency and Monetary is decided using the Elbow method. With the obtained clusters, it is possible to enrich the original data set and use this information as additional features. 'Topological RFM' also uses gradient tree boosting for doing prediction.
\section*{EXPERIMENTS} \label{sec:experiments}
To validate the two proposed methods in this work, 'Time Series RFM' and 'Topological RFM', this study carried out an assessment consisting of four settings: (1) Prediction without RFM, (2) RFM + Prediction, (3) Time Series RFM, (4) Topological RFM. As previously mentioned, gradient tree boosting was used as a predictive model. In this case, the implementation catboost\footnote{\url{https://catboost.ai}} was chosen. The reasoning behind is the touted support for categorical features. In this work, the obtained clusters are handled as categories. The data was divided into a training and test set using a 70-30 split.
To compare the quality of the results, Root Mean Square Error (RMSE) was used. This is defined as $RMSE = \sqrt{\frac{\sum_{t=1}^T (x_{1,t} - x_{2,t})^2}{T}}$. The results of the experiment can be found in \autoref{tab:models:1}.

  \begin{table}
\caption{Overview of results using mean RMSE. Low values are better. TS: Time Series. TDA: Topological Data Analysis}
\begin{center}\label{tab:models:1}
\footnotesize
\begin{tabulary}{\linewidth}{CCCCCCCCC}
    \toprule
Dataset	&	Model	&	RMSE	&	Dataset	&	Model	&	RMSE	&	Dataset	&	Model	&	RMSE	\\	\hline
CDNow	&	No RFM	&	13	&	Bimbo	&	No RFM	&	97	&	Cloud	&	No RFM	&	3.56	\\	
CDNow	&	RFM	&	12.56	&	Bimbo	&	RFM	&	172	&	Cloud	&	RFM	&	3.98	\\	
CDNow	&	\textbf{TS RFM}	&	3	&	Bimbo	&	TS RFM	&	291	&	Cloud	&	TS RFM	&	0.05	\\	
CDNow	&	TDA RFM	&	18.87	&	Bimbo	&	\textbf{TDA RFM}	&	18.97	&	Cloud	&	\textbf{TDA RFM}	&	0.03	\\	
Hospitality	&	No RFM	&	219	&	Hospitality	&	\textbf{TS RFM}	&	155	&		&		&		\\	
Hospitality	&	RFM	&	240	&	Hospitality	&	TDA RFM	&	275	&		&		&		\\	
 \bottomrule
 \end{tabulary}
 \end{center}
 \end{table}

\section*{DISCUSSION AND CONCLUSION}\label{sec:discussion}
This work proposes two methods driven by the idea that the recency, frequency and monetary aspects of a customer relation evolve over time and can be thus constructed as a time series. From a qualitative perspective, the benefit of using them is their highly visual component. The marketing analyst can show the results to decisions-makers. For example, by clustering the time series, it is possible to obtain a centroid, which is also a time series. The centroid provides important information on the behavior of the cluster. This can be communicated to the organization and help in the creation of 'customer personas'. Thus, analysts can continue using familiar tools and concepts and extend them. From a quantitative side, using either 'Time Series RFM' or 'Topological RFM' improves model accuracy. Interestingly, using RFM exclusively showed only minor improvements over not using RFM. Given the choice, it is better for the practitioner to avoid RFM. Another benefit of using 'Time Series RFM' or 'Topological RFM' is the improvement on accuracy of a machine learning model. Thus, they open the door for a 'CRM predictive pipeline', where the practitioner can segment her customer base, generate personas, do predictions and communicate to management both the predictions for the personas as well as for the individual users and identify those users diverging from the expected results from their respective personas.
As a next step, this work will seek to implement consensus clustering or clustering ensemble to avoid handling the three clusters of RFM as categorical variables, but rather merge them into a 'super cluster'.  Another important step is to identify the type of setting when it is a better option to use 'Topological RFM' over 'Time Series RFM' both from a theoretical perspective as well as from a practical one. It is important to generate heuristics that can guide quantitative analysts in their choice. Another direction is to take this line of work and expand it to very large data sets. For example, using the work of \cite{Lacombe2018-od} combining TDA with Optimal Transport to speed up the computation of persistence diagrams. Overall, TDA is a nascent field and to the best of the knowledge of this study, this is the first work dedicated to applying this techniques on CRM data to evaluate customer loyalty. As the field grows in popularity and new applications in marketing appear, it is to be expected that TDA will become an essential tool for the marketing practitioner.

\bibliographystyle{alpha}
\bibliography{IMP}

\newcommand{\etalchar}[1]{$^{#1}$}
\begin{thebibliography}{ZKNMK16}

\bibitem[AC18]{2018esade}
N.~{Agell} and M.~{Carricano}.
\newblock {Adopcion e impacto del Big Data y Advanced Analytics en Espana}.
\newblock {\em ESADE Business and Law School}, May 2018.

\bibitem[Aks13]{doi:10.1108/JOSM-01-2013-0018}
Lerzan Aksoy.
\newblock How do you measure what you can't define?: The current state of
  loyalty measurement and management.
\newblock {\em Journal of Service Management}, 24(4):356--381, 2013.

\bibitem[ASY15]{Aghabozorgi2015}
Saeed Aghabozorgi, Ali {Seyed Shirkhorshidi}, and Teh {Ying Wah}.
\newblock {Time-series clustering - A decade review}.
\newblock {\em Information Systems}, 53:16--38, 2015.

\bibitem[Cha13]{Chase2013}
Charles~W. Chase.
\newblock {\em {Demand-Driven Forecasting}}.
\newblock John Wiley {\&} Sons, Inc., Hoboken, NJ, USA, aug 2013.

\bibitem[CM17]{chazal2017introduction}
Fr{\'e}d{\'e}ric Chazal and Bertrand Michel.
\newblock An introduction to topological data analysis: fundamental and
  practical aspects for data scientists.
\newblock 2017.

\bibitem[FT05]{Fleisch2003}
Elgar Fleisch and Christian Tellkamp.
\newblock {Inventory inaccuracy and supply chain performance: a simulation
  study of a retail supply chain}.
\newblock {\em International Journal of Production Economics}, 95(3):373--385,
  mar 2005.

\bibitem[GHH{\etalchar{+}}06]{doi:10.1177/1094670506293810}
Sunil Gupta, Dominique Hanssens, Bruce Hardie, Wiliam Kahn, V.~Kumar, Nathaniel
  Lin, Nalini Ravishanker, and S.~Sriram.
\newblock Modeling customer lifetime value.
\newblock {\em Journal of Service Research}, 9(2):139--155, 2006.

\bibitem[GK17]{Gidea2017-em}
Marian Gidea and Yuri Katz.
\newblock Topological data analysis of financial time series: Landscapes of
  crashes.
\newblock March 2017.

\bibitem[LCO18]{Lacombe2018-od}
Th{\'e}o Lacombe, Marco Cuturi, and Steve Oudot.
\newblock Large scale computation of means and clusters for persistence
  diagrams using optimal transport.
\newblock May 2018.

\bibitem[PB17]{2017apec}
C.~{Pompa} and T.~{Burke}.
\newblock {Data Science and Analytics Skills Shortage: Equipping the APEC
  Workforce with the Competencies Demanded by Employers}.
\newblock {\em APEC Human Resource Development Working Group}, 2017.

\bibitem[PG17]{Paparrizos2017-nw}
John Paparrizos and Luis Gravano.
\newblock Fast and accurate {Time-Series} clustering.
\newblock {\em ACM Transactions on Database Systems (TODS)}, 42(2):8, June
  2017.

\bibitem[PR16]{Platzer2016-si}
Michael Platzer and Thomas Reutterer.
\newblock Ticking away the moments: Timing regularity helps to better predict
  customer activity.
\newblock {\em Marketing Science}, 35(5):779--799, September 2016.

\bibitem[RB17]{2017arXiv170905548R}
R.~{Rivera} and E.~{Burnaev}.
\newblock {Forecasting of commercial sales with large scale Gaussian
  Processes}.
\newblock {\em 17th International Conference on Data Mining Workshops (ICDMW),
  IEEE Conference Publications}, September 2017.

\bibitem[RNB18]{Rivera2018}
Rodrigo Rivera, Ivan Nazarov, and Evgeny Burnaev.
\newblock Towards forecast techniques for business analysts of large commercial
  data sets using matrix factorization methods.
\newblock {\em Journal of Physics: Conference Series}, 1117:012010, nov 2018.

\bibitem[SDB16]{Seversky2016-dz}
L~M Seversky, S~Davis, and M~Berger.
\newblock On {Time-Series} topological data analysis: New data and
  opportunities.
\newblock In {\em 2016 {IEEE} Conference on Computer Vision and Pattern
  Recognition Workshops ({CVPRW})}, June 2016.

\bibitem[TS19]{Turner2019-bc}
Katharine Turner and Gard Spreemann.
\newblock Same but different: distance correlations between topological
  summaries.
\newblock March 2019.

\bibitem[W{\"u}b08]{Wubben2008-yf}
Markus W{\"u}bben.
\newblock {\em Analytical {CRM}: Developing and Maintaining Profitable Customer
  Relationships in {Non-Contractual} Settings}.
\newblock Gabler Verlag, October 2008.

\bibitem[ZBS15]{Zhang2015-wa}
Yao Zhang, Eric~T Bradlow, and Dylan~S Small.
\newblock Predicting customer value using clumpiness: From {RFM} to {RFMC}.
\newblock {\em Marketing Science}, 34(2):195--208, March 2015.

\bibitem[ZKNMK16]{Zaki2016-oj}
Mohamed Zaki, Dalia Kandeil, Andy Neely, and Janet~R McColl-Kennedy.
\newblock The fallacy of the net promoter score: Customer loyalty predictive
  model.
\newblock 2016.

\end{thebibliography}

\end{document}